\documentclass[letterpaper, 10 pt, conference]{ieeeconf} 

\IEEEoverridecommandlockouts

\usepackage{epsfig} 
\usepackage{times} 
\usepackage[cmex10]{amsmath} 
\usepackage{amssymb}  

\usepackage{amsfonts}
\usepackage{graphicx}
\usepackage{algpseudocode}
\usepackage{algorithmicx}
\usepackage[caption=false,font=footnotesize]{subfig}
\usepackage{color}
\usepackage{cases}
\usepackage{bm}
\usepackage{cite}

\usepackage{fontawesome}

\algblock{ParFor}{EndParFor}
\algnewcommand\algorithmicparfor{\textbf{parallel for}}
\algnewcommand\algorithmicpardo{\textbf{do}}
\algnewcommand\algorithmicendparfor{\textbf{end\ parallel for}}
\algrenewtext{ParFor}[1]{\algorithmicparfor\ #1\ \algorithmicpardo}
\algrenewtext{EndParFor}{\algorithmicendparfor}

\newcommand{\rbb}{\mathbb{R}}

\newcommand{\p}[3][]{p\ifx\relax#1\relax\else_{#1}\fi\ifx\relax#2\relax\else^{#2}\fi\ifx\relax#3\relax\else\!\left( #3 \right)\fi}
\newcommand{\hp}[3][]{\widehat{p}\ifx\relax#1\relax\else_{#1}\fi\ifx\relax#2\relax\else^{#2}\fi\ifx\relax#3\relax\else\!\left( #3 \right)\fi}

\title{\LARGE\bf{Markerless visual servoing on unknown objects\\for humanoid robot platforms}}

\author{Claudio Fantacci$^{1}$, Giulia Vezzani$^{1}$, Ugo Pattacini$^{1}$, Vadim Tikhanoff$^{1}$ and Lorenzo Natale$^{1}$\thanks{$^{1}$Claudio Fantacci, Giulia Vezzani, Ugo Pattacini, Vadim Tikhanoff and Lorenzo Natale are with Istituto Italiano di Tecnologia, iCub Facility, Humanoid Sensing and Perception, Via Morego 30, Genova, Italy {\tt\small claudio.fantacci@iit.it}, {\tt\small giulia.vezzani@iit.it}, {\tt\small ugo.pattacini@iit.it}, {\tt\small vadim.tikhanoff@iit.it}, {\tt\small lorenzo.natale@iit.it}}}

\begin{document}
\maketitle
\thispagestyle{empty}
\pagestyle{empty}


\begin{abstract}
To precisely reach for an object with a humanoid robot, it is of central importance to have good knowledge of both end-effector, object pose and shape.
In this work we propose a framework for markerless visual servoing on unknown objects, which is divided in four main parts:
$\left.\mbox{\textsc{i}}\right)$ a least-squares minimization problem is formulated to find the volume of the object graspable by the robot's hand using its stereo vision;
$\left.\mbox{\textsc{ii}}\right)$ a recursive Bayesian filtering technique, based on Sequential Monte Carlo (SMC) filtering, estimates the 6D pose (position and orientation) of the robot's end-effector without the use of markers;
$\left.\mbox{\textsc{iii}}\right)$ a nonlinear constrained optimization problem is formulated to compute the desired graspable pose about the object;
$\left.\mbox{\textsc{iv}}\right)$ an image-based visual servo control commands the robot's end-effector toward the desired pose.
We demonstrate effectiveness and robustness of our approach with extensive experiments on the iCub humanoid robot platform, achieving real-time computation, smooth trajectories and sub-pixel precisions.
\end{abstract}


\section{INTRODUCTION}
Recent surge of interest in humanoid robots and their use in private or public contexts has risen the need for robust and resilient techniques for manipulation and interaction tasks.
These contexts present real-world challenges in that the environment is unstructured, complex and time varying.
Precise and reliable manipulation and interaction tasks can be achieved when accurate knowledge of both the object to manipulate and the end-effector pose is available.
This is possible for industrial settings, where it is required to repeat similar tasks over time, in a fine-calibrated setting and in a well-known and structured environment. 
Humanoid robots instead: $\left.\mbox{\textsc{i}}\right)$ are supposed to act in dynamic and unknown environment wherein object poses and shapes are unknown and
$\left.\mbox{\textsc{ii}}\right)$ have unreliable proprioception due to measurement noises, sensor biases, mechanical elasticity of the links and so forth.

In this paper, we propose a robust and reliable framework to address \textsc{i}, \textsc{ii} in the context of grasping tasks. In particular, we use vision both for modeling objects and their grasping poses, and compensating for the robot's proprioception errors. As a result, such a refined information allows designing of a \textit{visual servoing control} \cite{esp1992,hut1996,mal1999,kra2002,cha2006,cha2007} for precise reaching and grasping. Our approach is markerless and makes use of stereo vision information and RGB images.

Specifically, this work integrates our previous results \cite{Vezzani2017} and \cite{Fantacci2017} that respectively estimate the model and the grasping pose of an object with superquadric functions and the 6D pose of the robot end-effector by means of a particle filter. The combined exploitation of these approaches provide all the required input for addressing visual servoing problems.

As main contribution, we propose an image-based visual servoing approach with decoupled translation and orientation controls.
In particular, we formulate two different visual servoing problems.
In the first one we solve for the translation motion assuming the rotation is already completed.
Conversely, the latter computes the rotation motion under the assumption that the translation part is achieved.
Furthermore, we present practical solutions to use the particle filter estimates with visual servoing and a gain-scheduling technique to prevent the end-effector overshooting and oscillating around the goal pose.
Finally, we demonstrate the effectiveness of the proposed framework via experimental tests carried out in real-time on the iCub humanoid robot platform \cite{met2010}.

The rest of the paper is organized as follows.
Section \ref{sec:rel} reviews the state-of-art on visual servo control.
Section \ref{sec:background} briefly introduces the superquadric modeling and grasping pose computation, and the particle filter formulation.
Section \ref{sec:visualservo} gives details of the proposed image-based visual servo control.
In Section \ref{sec:exp} we report on the experiments to validate our approach.
Finally, Section \ref{sec:con} provides concluding remarks and future work.


\section{RELATED WORK}
\label{sec:rel}
The use of computer vision to control a robot manipulator motion toward an object, i.e. \textit{visual servoing} or \textit{visual servo control}, has been a well known research topic in literature for over two decades \cite{esp1992,hut1996,kra2002,cha2006,cha2007}.
The recent interest and development of humanoid robots have shifted the attention toward the integration of such methodologies to different humanoid robot platforms.

In \cite{Taylor2001}, the authors introduce a position-based visual servoing framework to overcome the problem of hand-eye calibration with an Extended Kalman Filter  tracking flashing LEDs on the robot hand. This, however, poses the constraint of using markers to track the end-effector and it requires markers to be visible during the whole motion.
The authors further extend their work by endowing their robot with an RGBD sensor \cite{Taylor2002}. Colour and depth information are used to fit box models onto target objects and then to estimate a graspable pose.

Another interesting technique to visual servoing is to use machine learning to estimate either, or both, the forward kinematics, relating the configuration of the arm joints with the position of the hand, along with the image Jacobian \cite{Hosoda1994,Lapreste2004,Sun2005,Natale2007}.

In \cite{Vahrenkamp2008,Huebner2009}, a hybrid visual servoing is used to grasp known objects on the humanoid robot ARMAR III. The proposed methodology uses prior information about the shape of the objects as well as a marker to track the robot's hand, which for generic contexts may be an unfeasible assumption.

In \cite{Gratal2011} the authors describe a visual servoing framework for grasping that is divided in several parts, two of which are for constructing the scene model and to estimate the pose of the end-effector.
Point clouds are used for scene reconstruction, while the Virtual Visual Servoing (VVS) \cite{esp2002,com2006} approach is used to estimate the pose of the robot's hand.
In particular, VVS uses a 3D rendering engine to virtually create 3D CAD models of the robot's end-effector as if it had been seen by the robot's camera.
Then, a classical visual servoing approach is used to move the virtual camera to overlap the rendered model of the end-effector with the real one in the image. The features used during visual servoing are provided by the Chamfer distance transform \cite{bor1988}.

In \cite{Vicente2017}, a Sequential Monte Carlo (SMC) algorithm estimates the offset present in the robot's encoders to correct the errors in the forward kinematics.
A simulator is used to generate predictions about hand appearance in the robot camera image plane.
These images are used to evaluate the likelihood by comparing RGB images using the Chamfer distance transform.

In the context of the present work, our framework uses stereo vision information to estimate the 3D shape and grasping poses of unknown objects. The robot's proprioception is refined using SMC filtering, which is robust and accounts for multimodal distributions. HOG descriptors are used to extract information about the end-effector shape from images without the use of markers. 


\section{PROPOSED FRAMEWORK}
\label{sec:graspapproach}
\begin{figure}[thpb]
\centering
	\framebox{\parbox{0.95\linewidth}{\includegraphics[width=\linewidth]{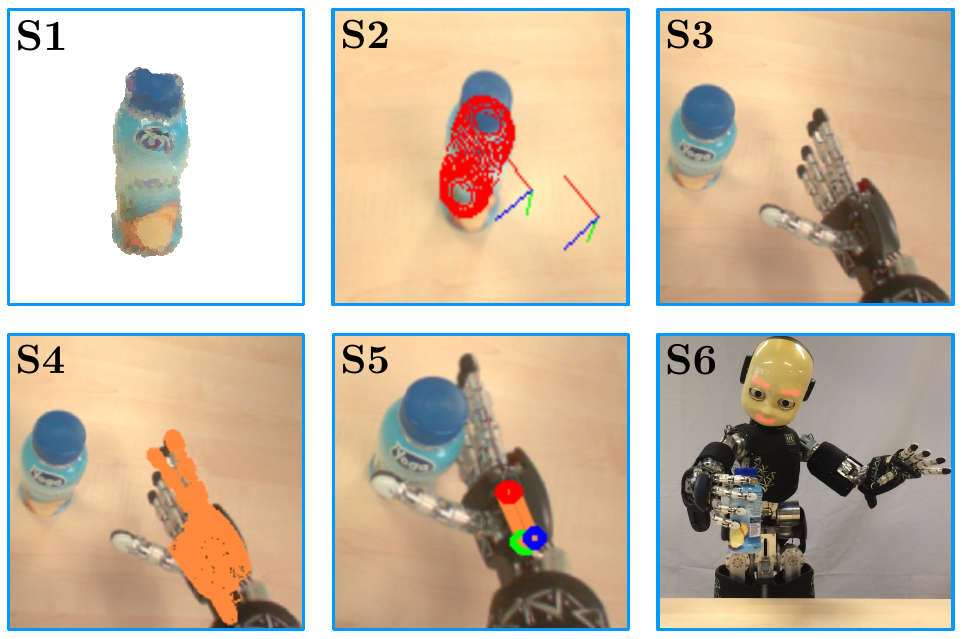}}}
        \caption{Block representation of the proposed markerless visual servoing framework on unknown objects.}
        \label{fig:pipeline}
\end{figure}

The framework we propose for markerless visual servoing  on unknown objects consists of the following steps (cfr. Fig. \ref{fig:pipeline}):
\begin{itemize}
	\item[\textbf{S1.}] The modeling approach described in~\cite{Vezzani2017} reconstructs a superquadric representing the object by using a 3D partial point cloud acquired from stereo vision.
	\item[\textbf{S2.}] The estimated model is exploited by the pose computation method of~\cite{Vezzani2017} for providing a grasping pose.
	\item[\textbf{S3.}] An open loop phase brings the robot's end-effector in the proximity of the object and in the cameras field-of-views.
	\item[\textbf{S4.}] The 3D model-aided particle filter of~\cite{Fantacci2017} estimates the end-effector pose using RGB images.
	\item[\textbf{S5.}] Visual servoing uses the particle filter output of \textbf{S4.} in order to reach for the pose computed in \textbf{S2.}.
	\item[\textbf{S6.}] Reaching completes and the robot grasps the object.
\end{itemize}

For the sake of completeness Section \ref{sec:background} reports \textbf{S1}-\textbf{S4}, whereas Section \ref{sec:visualservo} details the design of the visual servo control.

\section{BACKGROUND}
\label{sec:background}
In this Section, we  provide the background on the grasping pose computation of \cite{Vezzani2017} and on 3D model-aided particle filtering of \cite{Fantacci2017}.

\subsection{Grasping approach}
\label{ssec:posecomp}
The grasping approach used in this work is based on modeling the object and the volume graspable by the robot hand with superquadric functions.

Superquadrics are a generalization of quadric surfaces and includes supertoroids, superhyperboloids and superellipsoids.  In this work, we focus on \textit{superellipsoids} -- that we will call simply superquadrics from now on -- since they identify closed surfaces and, thus, are suitable for object modeling.

A superquadric can be represented in an object-centered system with the \textit{inside-outside} function:
\begin{equation}
	F(x,y,z, \bm{\lambda}) = \left( \left( \frac{x}{\lambda_1}\right) ^{\frac{2}{\lambda_5}} + \left( \frac{y}{\lambda_2}\right) ^{\frac{2}{\lambda_5}}\right) ^{\frac{\lambda_5}{\lambda_4}}+ \left( \frac{z}{\lambda_3}\right) ^{\frac{2}{\lambda_4}}\,,
	\label{e:superq}
\end{equation}
where the five parameters $\bm{\lambda}=[\lambda_1, \dots, \lambda_5]$ defines the superquadric dimensions and shape. Equation \eqref{e:superq}  provides a simple test whether a given point lies ($F = 1$) or not ($F > 1$ or $F < 1$) on the superquadric surface.

In the next paragraphs, we briefly recall the superquadric modeling and the grasping pose computation proposed in \cite{Vezzani2017}.

\subsubsection{\textbf{Superquadric modeling}}
\label{sssec:superquadricmodeling}
Object modeling with superquadrics consists of finding that superquadric $\mathcal{O}$ which best represents the object  by using a single and partial 3D point cloud, acquired by stereo vision. In particular, we need to estimate those values of the parameters vector $\bm{\lambda} \in \mathbb{R}^{11}$, so that most of the $N$ 3D points $\bm{m}_{i}^{o}=\left[x_i, y_i, z_i \right] $ for $i=1, \dots, N$, collected from the object surface,  lie on or close to the superquadric.
The minimization of the algebraic distance from points to the model can be solved by defining a least-squares minimization problem
\begin{equation}
	\min_{\bm{\lambda}} \sum_{i=1}^{N}\left(\sqrt{\lambda_1 \lambda_2 \lambda_3} \left(F(\bm{m}_{i}^{o},\bm{\lambda})-1 \right) \right)^2 \,, 
	\label{e:objrec}
\end{equation}
where $\left(F(\bm{m}_{i}^{o},\bm{\lambda})-1 \right) ^2$ imposes the point-superquadric distance minimization and the term $\lambda_1 \lambda_2 \lambda_3$, which is proportional to the superquadric volume, compensates for the fact that the previous equation is biased towards larger superquadric.

The optimization problem of Equation \eqref{e:objrec} is solved in real-time by Ipopt \cite{ipopt}, a software package for large scale nonlinear constrained optimization problem.

\begin{figure}[thpb]
	\centering
	\subfloat[]
	{
		\framebox{\parbox{0.3\linewidth}{\includegraphics[width=\linewidth]{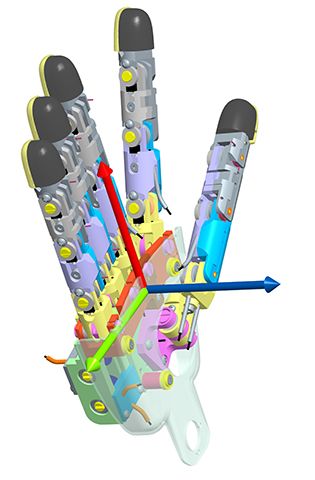}}}
		\label{handframe}		
	}
	\subfloat[]
	{
		\framebox{\parbox{0.346\linewidth}{\includegraphics[width=\linewidth]{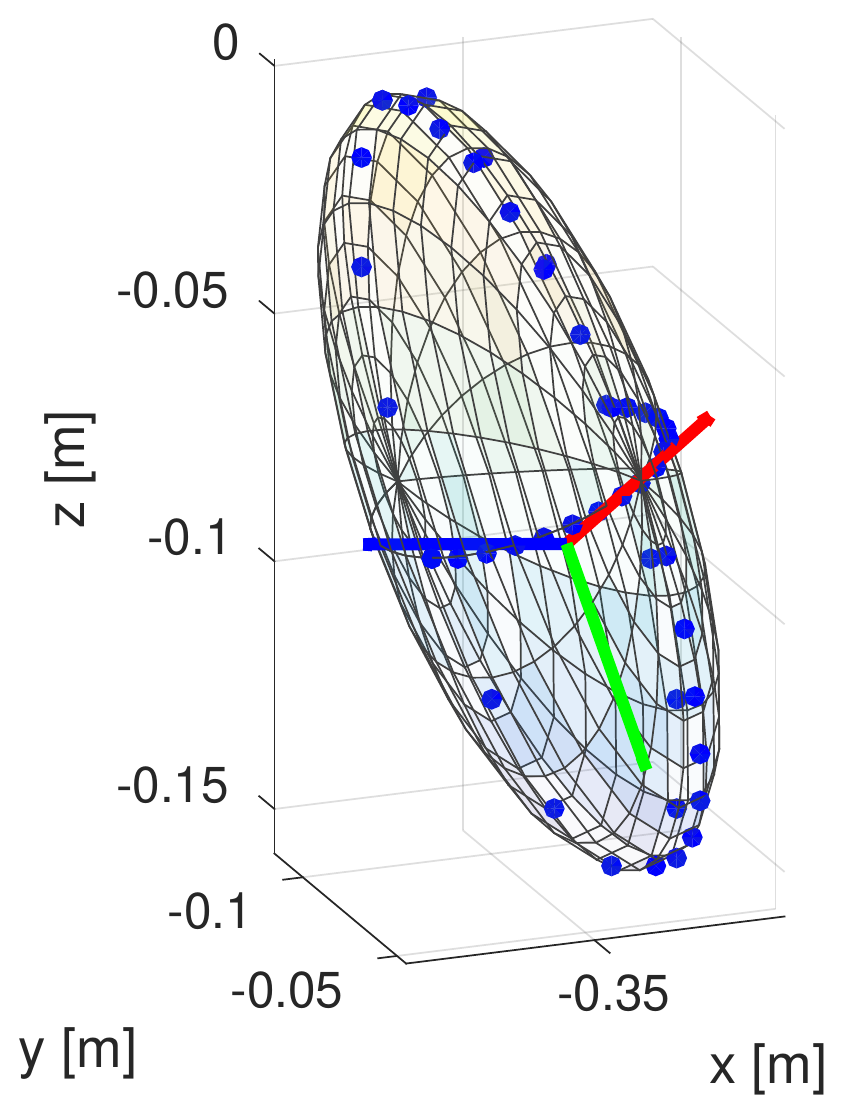}}}
		\label{points}
	}
	\caption{In Fig. (a),  the reference frame $(\bm{x^{g}}, \bm{y^{g}}, \bm{z^{g}})$ attached to the robot hand in RGB convention ($\bm{x^{g}}$ is coloured in red, $\bm{y^{g}}$ in green, $\bm{z^{g}}$ in blue). In Fig (b): the $L$ points sampled on the closest half of the hand ellipsoid $\mathcal{H}$. The RGB frame represents the hand pose, showing how the ellipsoid $\mathcal{H}$ is attached to the hand.} \label{fig:points}
\end{figure}

We use a superquadric function for representing also the volume graspable by the robot's hand. In this case, the shape and pose of such superquadric are known a-priori, as they depend on  the hand shape  and its grasping capabilities. A suitable shape for this purpose turns out to be the ellipsoid $\mathcal{H}$ attached to the hand palm shown in Fig. \ref{fig:points}.

\subsubsection{\textbf{Grasping pose computation}}
\label{sssec:graspingpose}
The approach described in~\cite{Vezzani2017} provides a feasible grasping pose for the robot hand by using the object superquadric $\mathcal{O}$ and the ellipsoid  $\mathcal{H}$ modeling the volume graspable by the hand. The hand pose is represented with a 6D vector $\bm{x}^{g}=[x^{g}, y^{g}, z^{g}, \phi^{g}, \theta^{g}, \psi^{g}]^{\top}$, where $(x^{g}, y^{g}, z^{g})$ are the coordinates of the origin of the hand frame and $(\phi^{g}, \theta^{g}, \psi^{g})$ are the RPY Euler angles, accounting for orientation.

The basic concept of this grasping approach is  to compute the solution by looking for a pose  $\bm{x}^{g}$ that makes the hand ellipsoid $\mathcal{H}$ overlap with the object superquadric $\mathcal{O}$ while meeting a set of requirements that guarantee $\bm{x}^{g}$ is reachable by the robot's hand.

The general formulation of the problem can be described by the following nonlinear constrained optimization:
\begin{IEEEeqnarray}{rl}
		\min_{\bm{x}^{g}} & \sum_{i=1}^{L}\left( \sqrt{\lambda_1 \lambda_2 \lambda_3} \left(F(\bm{m}_i^{\bm{x}^{g}},\bm{\lambda})-1 \right) \right)^2 \nonumber\\
		\mbox{subject to:} & 	\label{e:grasp}\\
		& h_i(\bm{a_i}, c_{i}(\bm{m}_1^{\bm{x}^{g}}, \dots, \bm{m}_L^{\bm{x}^{g}})) >0 \nonumber\\ 
		& \mbox{for } i = 1, \dots, M \, . \nonumber
\end{IEEEeqnarray}
Hereafter, we briefly recall the meaning of the most
important quantities of Eq. \eqref{e:grasp}. An exhaustive description of the pose computation approach is provided in~\cite{Vezzani2017}.
\begin{itemize}
	\item \textit{The cost function} imposes the minimization of the distance between the object superquadric $\mathcal{O}$, represented by the inside-outside function $\left(F(\cdot,\bm{\lambda})-1\right)$, and $L$ points $\bm{m}_i^{\bm{x}^{g}}=\left[ p_{x,i}^{\bm{x}^{g}},p_{y,i}^{\bm{x}^{g}},p_{z,i}^{\bm{x}^{g}}\right] $ for $i=1, \dots, L$, properly sampled on the surface of the  hand ellipsoid $\mathcal{H}$, whose pose is given by vector $\bm{x}^{g}$.
	\item The $M$ \textit{constraints} of Eq.~\eqref{e:grasp} take into account obstacle avoidance requirements. Each term $h_i$, for  $i=1, \dots, M$ is the implicit function representing the $i$-th  obstacle. As is in \cite{Vezzani2017}, the only obstacle of our scenario is the table on which the object is located, hence $M = 1$. The quantity $h_1(\bm{a}_1, c_1(\cdot))$ = $h(\bm{a}, c(\cdot))$ is the implicit function of the plane modeling the table. The vector $\bm{a}$ consists of the parameters of the plane function and each $f(\bm{m}_1^{\bm{x}^{g}}, \dots, \bm{m}_L^{\bm{x}^{g}})$ accounts for a dependency on the $L$ points $\bm{m}_i$ suitably designed for the grasping task.
\end{itemize}
 
We solve the optimization problem of Equation \eqref{e:grasp} with the Ipopt package efficiently and with execution times compatible with online applications (nearly 2 seconds on average).
 
\subsection{3D model-aided particle filtering}
\label{ssec:3dsispf}
The objective of a SMC filter, or \textit{particle filter} (PF), is to provide a numerical solution to the recursive Bayesian filter defined by the the Chapman-Kolmogorov equation and the Bayes' rule \cite{ho1964}:
\begin{IEEEeqnarray}{rCl}
	\p[k|k-1]{}{\bm{x}} & = & \int \varphi_{k|k-1}\!\left( \bm{x} | \bm{\zeta} \right) \p[k-1]{}{\bm{\zeta}} d \bm{\zeta} \, ,\label{eq:chapkol}\\
	\p[k]{}{\bm{x}} & = & \dfrac{g_{k}\!\left( \bm{y}_{k} | \bm{x} \right) \, \p[k|k-1]{}{\bm{x}}}{\displaystyle \int g_{k}\!\left( \bm{y}_{k} | \bm{\zeta} \right) \, \p[k|k-1]{}{\bm{\zeta}} d \bm{\zeta}} \, ,\label{eq:bayesrule}
\end{IEEEeqnarray}
where $k$ is the time instant, $\bm{y}_{k} \in \rbb^{n_{y}}$ is the noisy measurement at time $k$, $\p[k|k-1]{}{\cdot}$ is the \textit{predicted density}, $\p[k]{}{\cdot}$ is the \textit{posterior density} and $\varphi_{k+1|k}\!\left( \cdot | \zeta \right)$ is a \textit{Markov transition density} and $g_{k}\!\left( y | \cdot \right)$ is the \textit{measurement likelihood function} \cite{gor1993,dou2000,dou2001,aru2002,ris2004}.

Particle filters characterize the \textit{posterior density} $\p[k]{}{\bm{x}}$ of the state $\bm{x}_{k} \in \rbb^{n_{x}}$ of a dynamical system at time $k$, with a set of support points, termed \textit{particles}, and associated weights $\{ \bm{x}^{(i)}_{k},\, w^{(i)}_{k} \}_{1\leq i \leq N}$.
The principle under which a particle set is able to approximate the posterior density $\p[k]{}{\bm{x}}$ over time is the \textit{importance sampling} \cite{dou2000,rob2004}.
During importance sampling, a \textit{proposal} or \textit{importance density} $\pi_{k}\!\left( \bm{x}_{k} | \bm{x}_{k-1}, \, \bm{y}_{k} \right)$ is used to draw preliminary particles at time $k$:
\begin{equation}
	\bm{x}^{(i)}_k  \sim  \pi_{k}\!\left( \bm{x} | \bm{x}^{(i)}_{k-1}, \, \bm{y}_k \right) \, , \label{e:pf1}
\end{equation}
whose weights are computed as follows:
\begin{IEEEeqnarray}{rCl}
	\widetilde{w}^{(i)}_{k} & = & w^{(i)}_{k-1}
	\frac{g_{k}(\bm{y}_{k} | \bm{x}^{(i)}_{k})\,\varphi_{k | k-1}(\bm{x}^{(i)}_{k} | \bm{x}^{(i)}_{k-1})}{\pi_{k}(\bm{x}^{(i)}_{k} | \bm{x}^{(i)}_{k-1}, \bm{y}_{k})} \label{e:unnw}\\
	w^{(i)}_{k} & = &\frac{\widetilde{w}^{(i)}_{k}}{\sum_{j=1}^{N} \widetilde{w}^{(j)}_{k}} \label{e:pf3}
\end{IEEEeqnarray}
for $i=1,\dots,N$.
To avoid that the particle weights degenerate to a situation where all except few become zero, a \textit{resampling step} \cite{dou2000,dou2001,aru2002,ris2004}.
The described particle method is known in literature as Sequential Importance Sampling (SIS) \cite{aru2002,ris2004}.

The main advantages of particle filter methods are that they can deal with arbitrary nonlinearities and distributions (including multimodal ones), and can supply a complete representation of the posterior state distributions that improves as $N\rightarrow \infty$ \cite{gor1993,dou2000,dou2001,aru2002}.
The most suboptimal choice of the proposal is $\pi_{k} \equiv \varphi_{k|k-1}$, i.e. the transitional density \cite{ris2004}.

To adapt the SIS PF to our needs, we need to choose:
\begin{enumerate}
	\item the initialization procedure;
	\item the Markovian transition density defining the proposal;
	\item the likelihood model.
\end{enumerate}
For the sake of simplicity the following subsections consider a single camera viewpoint at time instant $k$.

\subsubsection{\textbf{Initialization}}
the state of the pose of the end-effector is denoted by $\bm{x} = \left[ p_{x}, p_{y}, p_{z}, u_{x}, u_{y}, u_{z}, \theta \right]^{\top}$, where:
$\left( p_{x}, \, p_{y}, \, p_{z}, \right)$ is the Cartesian position of the end-effector with respect to the root reference of the robot \cite{met2010};
$\left( u_{x}, u_{y}, u_{z}, \theta \right)$ is the orientation of the end-effector expressed in axis-angle notation.
We use the joint angles from the robot's base frame to the end-effector frame $\bm{q}^{e}_{k} = \left\{ q^{e}_{k,\,1}, \dots, q^{e}_{k,\,n} \right\}$ to set the particle $\bm{x}^{(i)}_{0}$, $1 \leq i \leq N$, equal to the pose of the hand provided by the \textit{direct kinematics} map $\kappa\!\left( \bm{q}^{e}_{k} \right)$ \cite{sic2016}.

\subsubsection{\textbf{Markovian transition density}}
we use the direct kinematics map $\kappa\!\left( \bm{q}^{e}_{k} \right)$ to model, given a certain motor command $\Delta\bm{q}^{e}_{k-1}$, the motion of the end-effector $\bm{x}_{k}$ subject to uncertainties and disturbances as
\begin{equation}
	\bm{x}_{k} = f_{k-1}\!\left( \bm{x}_{k-1}, \Delta\bm{q}^{e}_{k-1} \right) + \bm{w}_{k-1}
\end{equation}
where $\bm{w}_{k-1}$ is the process noise.
The Markov transition density results to be the described by the PDF
\begin{equation}
	\varphi_{k|k-1}\!\left( \bm{x}_{k} | \bm{x}_{k-1}, \Delta\bm{q}^{e}_{k-1} \right) \! = \! \p[w]{}{\bm{x}_{k} \! - \! \bm{f}_{k-1}\!\left( \bm{x}_{k-1}, \Delta\bm{q}^{e}_{k-1} \right)} \! . \label{eq:markovpred}
\end{equation}

\subsubsection{\textbf{Likelihood model}}
\begin{figure}[thpb]
\centering
	\framebox{\parbox{0.8\linewidth}{\includegraphics[width=\linewidth]{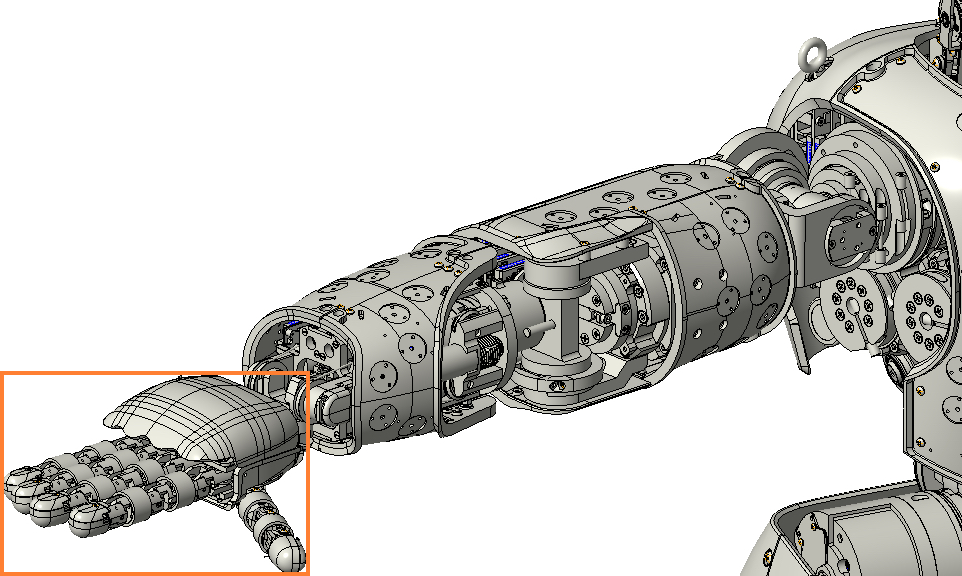}}}\vspace{0.3em}
	\framebox{\parbox{0.379\linewidth}{\includegraphics[width=\linewidth]{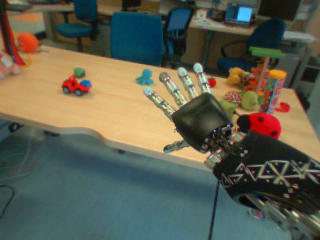}}}
	\framebox{\parbox{0.379\linewidth}{\includegraphics[width=\linewidth]{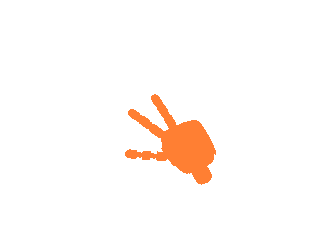}}}
        \caption{\textbf{Top:} mechanical model of the iCub right arm. \textbf{Bottom left:} image from the left camera of iCub. \textbf{Bottom right:} rendered image of the right end-effector (hand) of the iCub. In the context of this work, we decided to disable the ring and little fingers from being rendered. Motivations are detailed in Section \ref{sec:exp}.}
        \label{fig:render}
\end{figure}

given a pose $\bm{x}$, we use the OpenGL 3D rendering engine \cite{shr2013} to create a virtual image $\widehat{I}_{k}$ of the robot's end-effector as it would be seen by the current robot's camera point of view.
In particular, each particle $x_{k}^{(i)}$ represents a pose of the end-effector for which a virtual image
\begin{equation}
\widehat{I}_{k}^{(i)} \triangleq r\!\left( \bm{x}_{k}^{(i)}, \bm{q}^{c}_{k}, \bm{K} \right) \, , \label{eq:rendering}
\end{equation}
is rendered using the joint angles $\bm{q}^{c}_{k}$ of the camera kinematic chain and the \textit{intrinsic matrix} $\bm{K}$ \cite{har2003}.
Within (\ref{eq:rendering}), the projection matrix
\begin{equation}
	\bm{\Pi} = \bm{K} \, \bm{H}\!\left( \bm{q}^{c}_{k} \right) \label{eq:proj}
\end{equation}
performs a homogeneous transformation from the robot base reference frame to the camera image plane, with $\bm{H}\!\left( \bm{q}^{c}_{k} \right)$ the homogeneous transformation from the base reference frame to the camera frame.
A pictorial representation of the rendering process (\ref{eq:rendering}) is shown in Fig. \ref{fig:render}.

Using (\ref{eq:rendering}) and images from the robot's camera, we define the \textit{measurement likelihood function} $g_{k}\!\left( \cdot | \cdot \right)$ on the HOG descriptor $\bm{y}_{k}$ extracted from the camera image and the descriptor $\widehat{\bm{y}}^{(i)}_{k}$ extracted from the rendered image as follows:
\begin{IEEEeqnarray}{rCl}
	g_{k}\!\left( \bm{y}_{k} | \bm{x}^{(i)}_{k} \right) & \triangleq & e^{-\dfrac{1}{\sigma}\left| \bm{y}_{k} - \widehat{\bm{y}}^{(i)}_{k} \right|} \, ,\label{eq:hoglik}
\end{IEEEeqnarray}
where $\sigma$ is a free tuning parameter.

Further details and a pseudo code of the 3D model-aided SIS PF can be found in \cite{Fantacci2017}.

\subsection{State estimate extraction method}
\label{ssec:extraction}
After each PF cycle, a new estimation of the 6D pose of the hand is available.
However, such estimates have a non-smooth trajectory and, as a result, are not suitable for a control loop that should provide a smooth, straight and precise trajectory of the robot's end-effector.
To tackle this problem we use a \textit{Moving Average} (MA) technique to regularize the output of the PF.

\section{Visual servo control}
\label{sec:visualservo}
The goal of visual servoing is to command the robot's end-effector for accurately reaching a desired pose.
A good visual servoing approach for humanoid robots requires the design of a robust and reliable control law and a human-like motion of the upper-body. 
To this end, we set the following requirements:
\begin{itemize}
\setlength\itemsep{0.5em}
	\item[$\left( \textsc{i} \right)$] \textit{The image Jacobian shall provide a velocity screw of the end-effector in the Cartesian domain.} Instead of controlling each joint velocity, we command the end-effector trajectory using a \textit{Cartesian controller} \cite{pat2010}. The main advantage of using this approach is that it $\left.\mbox{\textsc{i}}\right)$ automatically deals with singularities;  $\left.\mbox{\textsc{ii}}\right)$ automatically accounts for joint limits; $\left.\mbox{\textsc{iii}}\right)$ can find solutions in virtually any working conditions \cite{pat2010}.
	\item[$\left( \textsc{ii} \right)$] \textit{The end-effector trajectory shall be as straight as possible.} This particular requirement is to simplify motion planning, which usually has an initial open loop phase to bring the end-effector in the proximity of the object to manipulate.
\end{itemize}

The two main ingredients to design a visual servo control are the goal pose
\begin{equation}
	\bm{x}^{g} = \left[ p_{x}^{g}, p_{y}^{g}, p_{z}^{g}, u_{x}^{g}, u_{y}^{g}, u_{z}^{g}, \theta^{g}\right]^{\top}
\end{equation}
and the current pose of the end-effector
\begin{equation}
	\bm{x}^{e} = \left[ p_{x}^{e}, p_{y}^{e}, p_{z}^{e}, u_{x}^{e}, u_{y}^{e}, u_{z}^{e}, \theta^{e}\right]^{\top} \,,
\end{equation}
where $\left( p_{x}^{*},\, p_{y}^{*},\, p_{z}^{*} \right)$ are the 3D Cartesian coordinates and $\left( u_{x}^{*},\, u_{y}^{*},\, u_{z}^{*},\, \theta^{*} \right)$ is the axis-angle representation of the orientation.
$\bm{x}^{g}$ is provided by the the grasping pose computation of Section \ref{ssec:posecomp}, while  $\bm{x}^{e}$ is given by the 3D model-aided particle filter of Section \ref{ssec:3dsispf}.
The visual servoing objective is to minimize the error
\begin{equation}
	\bm{e}_{k} \triangleq \bm{s}(\bm{x}^{e}_{k}) - \bm{s}(\bm{x}^{g}) = \bm{s}^{e}_{k} - \bm{s}^{g} \, , \label{eq:visualservoerror}
\end{equation}
where $\bm{s}^{e}_{k}$ and $\bm{s}^{g}$ are some feature representing, respectively, the manipulator and the goal pose.
Once a feature $\bm{s}$ is selected, the aim is to design a velocity controller.
To do so, \textit{considering the object stationary}, we require the relationship between the variation of $\bm{s}^{e}_{k} \in  \rbb^{m}$ and the end-effector velocity.
Denote the spatial velocity of the end-effector as
\begin{IEEEeqnarray}{rCl}
	\dot{\bm{x}}^{e} & \triangleq & \left[ \, \bm{v},\, \bm{\omega} \, \right]^{\top} \in \rbb^{6},\, \label{eq:velocityscrew}\\
	\bm{v} & = & \left[ \, v_{x},\, v_{y},\, v_{z} \, \right]^{\top} ,\, \\
	\bm{\omega} & = & \left[ \, \omega_{x},\, \omega_{y},\, \omega_{z} \, \right]^{\top} ,\,
\end{IEEEeqnarray}
with $\bm{v}$ the linear velocity of the origin and $\bm{\omega}$ the instantaneous angular velocity of the manipulator frame.
The relationship between $\dot{\bm{s}}^{e}_{k}$ and $\dot{\bm{x}}^{e}$ is described by the equation
\begin{equation}
	 \dot{\bm{s}}^{e}_{k} = \bm{J} \dot{\bm{x}}^{e} \, ,
\end{equation}
where $\bm{J} \in \rbb^{m \times 6}$ is the \textit{feature Jacobian}, or simply \textit{Jacobian}, and from which, using (\ref{eq:visualservoerror}), is possible to derive a control law with exponential decrease of the error of the form
\begin{equation}
	\dot{\bm{x}}^{e} = -K^{e} \bm{J}^{\dagger} e \, , \label{eq:controlvel}
\end{equation}
with $K^{e} > 0$ a proportional gain and $\bm{J}^{\dagger}$ the Moore-Penrose pseudo-inverse of the Jacobian.

The visual servoing approaches can be divided in two categories: \textit{image-based visual servoing} and \textit{position-based visual servoing} \cite{cha2006}.
The first approach uses image-plane coordinates of a set of points to define the feature vector $\bm{s}^{*}$, while the latter directly uses the pose for $\bm{s}^{*}$.
It would be natural, in our setting, to use a position-based visual servo control since we estimate both the pose of the goal and of the end-effector.
However, image-based visual servoing is preferable because it allows precise control, despite errors in the extrinsic camera parameters.

Image-based visual servo controls are known to be robust to camera and robot calibration errors \cite{hut1996}, but produce poorly predictable Cartesian trajectory \cite{mal1999}.
To tackle this problem, it is important to construct a good Jacobian $\bm{J}$ that $\left. 1 \right)$ ensures an exponential decrease of the error $\bm{e}$; $\left. 2 \right)$ guarantees that the trajectory of the feature points will follow a straight line from their initial to their desired positions; $\left. 3 \right)$ avoids unexpected translational motion when the rotation between the initial and desired configurations is large \cite{cha2006}.
In order to design a proper control law, let us introduce the image feature $\textbf{s}$ and the corresponding generic formulation of the image Jacobian $\bm{J}$.

\begin{figure}[thpb]
\centering
	\framebox{\parbox{0.95\linewidth}{\includegraphics[width=\linewidth]{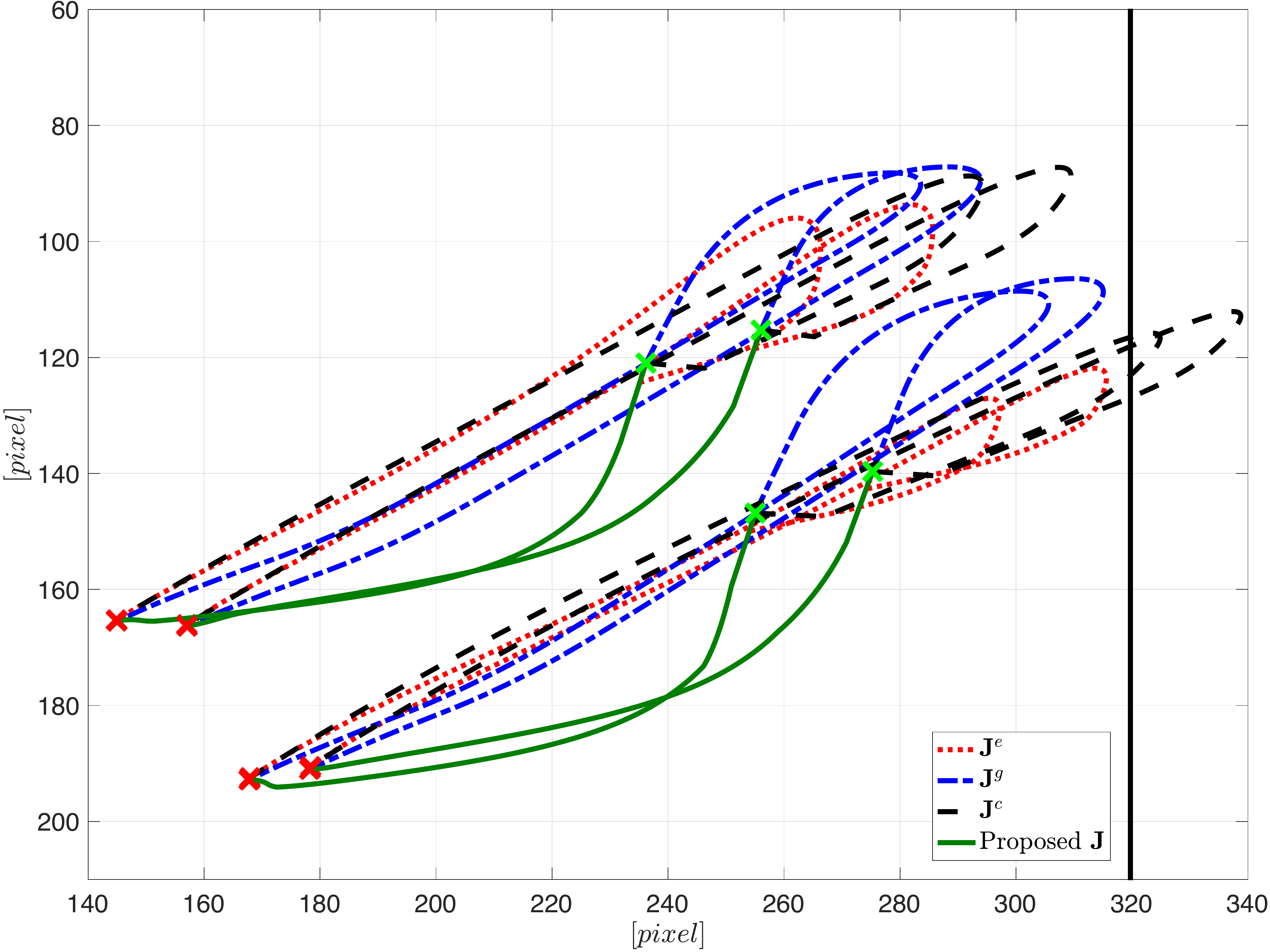}}}
        \caption{Left camera view of four image-plane trajectories performed by the right end-effector using different image Jacobians. The green and red crosses represent, respectively, the initial and final position of the end-effector. The reaching task was carried out in simulation to avoid damaging the robot, and it mainly consists of a translation toward the left and a small positive rotation. The solid black line on the right highlights the end of the image frame, which in our setting is $320 \times 240$. Note that only the green solid line, representing our image-based visual servoing control, is capable of providing a satisfactory trajectory.}
        \label{fig:trajectories}
\end{figure}

To control the 6D pose of the end-effector and to avoid configurations in which $\bm{J}$ becomes singular, four different visual features can be considered from both $\bm{x}^{g}$ and $\bm{x}^{e}$ \cite{cha2006}.
In particular, we define four coplanar 3D points around both $\bm{x}^{g}$ and $\bm{x}^{e}$ that are in turn projected on both left and right camera image plane with (\ref{eq:proj}), i.e.
\begin{equation}
	\bm{\Pi}_{l} \bm{x}^{e}_{i} = z^{e}_{l,\,i}
	\left[
		\begin{IEEEeqnarraybox*}[][c]{,c,}
			u^{e}_{l,\,i} \vspace{0.5em}\\
			v^{e}_{l,\,i} \vspace{0.5em} \\
			1
		\end{IEEEeqnarraybox*}
	\right]
	,\,\hspace{0.7em}
	\bm{\Pi}_{r} \bm{x}^{e}_{i} = z^{e}_{r,\,i}
	\left[
		\begin{IEEEeqnarraybox*}[][c]{,c,}
			u^{e}_{r,\,i}  \vspace{0.5em}\\
			v^{e}_{r,\,i}  \vspace{0.5em}\\
			1
		\end{IEEEeqnarraybox*}
	\right]
	\,,\hspace{0.7em}
	1 \leq i \leq 4 \,; \label{eq:featureend}
\end{equation}
\begin{equation}
	\bm{\Pi}_{l} \bm{x}^{g}_{i} = z^{g}_{l,\,i}
	\left[
		\begin{IEEEeqnarraybox*}[][c]{,c,}
			u^{g}_{l,\,i}  \vspace{0.5em}\\
			v^{g}_{l,\,i}  \vspace{0.5em}\\
			1
		\end{IEEEeqnarraybox*}
	\right]
	,\,\hspace{0.7em}
	\bm{\Pi}_{r} \bm{x}^{g}_{i} = z^{g}_{r,\,i}
	\left[
		\begin{IEEEeqnarraybox*}[][c]{,c,}
			u^{g}_{r,\,i}  \vspace{0.5em}\\
			v^{g}_{r,\,i}  \vspace{0.5em}\\
			1
		\end{IEEEeqnarraybox*}
	\right]
	\,,\hspace{0.7em}
	1 \leq i \leq 4 \,. \label{eq:featuregoal}
\end{equation}
In order to evaluate the error (\ref{eq:visualservoerror}), we define the visual feature $s$ as:
\begin{IEEEeqnarray}{rCl}
	\bm{s}^{e} & = & \Big[ \bar{\bm{s}}^{e}_{1},\, \bar{\bm{s}}^{e}_{2},\, \bar{\bm{s}}^{e}_{3},\, \bar{\bm{s}}^{e}_{4} \Big]^{\top} \label{eq:featee}\\
	\bar{\bm{s}}^{e}_{i} & \triangleq & \Big[ u^{e}_{l,\,i},\, u^{e}_{r,\,i},\, v^{e}_{l,\,i},\, v^{e}_{r,\,i} \Big] \,, \hspace{1.5em} 1 \leq i \leq 4\,, \vspace{0.5em}\\
	\bm{s}^{g} & = & \Big[ \bar{\bm{s}}^{g}_{1},\, \bar{\bm{s}}^{g}_{2},\, \bar{\bm{s}}^{g}_{3},\, \bar{\bm{s}}^{g}_{4} \Big]^{\top} \label{eq:featgoal} \\
	\bar{\bm{s}}^{g}_{i} & \triangleq & \Big[ u^{g}_{l,\,i},\, u^{g}_{r,\,i},\, v^{g}_{l,\,i},\, v^{g}_{r,\,i} \Big] \,, \hspace{1.5em} 1 \leq i \leq 4\,.
\end{IEEEeqnarray}
Finally, to relate changes in image point coordinates to changes in the Cartesian pose of the robot's manipulator, the general image Jacobian $\bm{J}$ is calculated as follows \cite{hut1996}:
\begin{IEEEeqnarray}{rCl}
	\bm{J} & = & \operatorname{\textsc{row}}\!\Big(
		\begin{IEEEeqnarraybox*}[][c]{,c,}
			\bm{J}_{1},\, \bm{J}_{2},\, \bm{J}_{3},\, \bm{J}_{4}
		\end{IEEEeqnarraybox*}
	\Big) \in \rbb^{16 \times 6}, \label{eq:jacobian}\\
	\bm{J}_{i} & = & \small{ \left [
	\begin{IEEEeqnarraybox*}[][c]{c;c;c;c;c;c}
		-\frac{f_{l}}{z_{l,\,i}} & 0 & \frac{u_{l,\, i}}{z_{l,\,i}} & \frac{u_{l,\, i}v_{l,\, i}}{f_{l}} & -\frac{f_{l}^{2} + \left( u_{l,\, i} \right)^{2}}{f_{l}} & v_{l,\, i} \vspace{0.5em}\\
		-\frac{f_{r}}{z_{r,\,i}} & 0 & \frac{u_{r,\, i}}{z_{r,\,i}} & \frac{u_{r,\, i}v_{r,\, i}}{f_{r}} & -\frac{f_{r}^{2} + \left( v_{r,\, i} \right)^{2}}{f_{r}} & v_{r,\, i} \vspace{0.5em}\\
		0 & -\frac{f_{l}}{z_{l,\,i}} & \frac{v_{l,\, i}}{z_{l,\,i}} & \frac{f_{l}^{2} + \left( v_{l,\, i} \right)^{2}}{f_{l}} & -\frac{u_{l,\, i}v_{l,\, i}}{f_{l}} & -u_{l,\, i} \vspace{0.5em}\\
		0 & -\frac{f_{l}}{z_{r,\,i}} & \frac{v_{r,\, i}}{z_{r,\,i}} & \frac{f_{r}^{2} + \left( v_{r,\, i} \right)^{2}}{f_{r}} & -\frac{u_{r,\, i}v_{r,\, i}}{f_{r}} & -u_{r,\, i}
	\end{IEEEeqnarraybox*}
	\right]},  \vspace{0.5em} \nonumber
\end{IEEEeqnarray}
where $f_{l}$ and $f_{r}$ are, respectively, the \textit{focal length} of the left and right camera, which are known from the camera calibration matrix $K$ in (\ref{eq:proj}), and $\operatorname{\textsc{row}}\!\left( \cdot \right)$ is a row-wise matrix stacking operator.
Note that the image coordinates $\left( u_{*}, v_{*} \right)$ do not specify any of the two possible superscript $e$ or $g$.
This is because several choice are available, each generating a different velocity screw $\dot{\bm{x}}^{e}$.
Popular approaches are \cite{cha2006}:
\begin{enumerate}
	\item evaluate $\bm{J} \triangleq \bm{J}^{e}$ using (\ref{eq:featureend});
	\item evaluate $\bm{J} \triangleq \bm{J}^{g}$ using (\ref{eq:featuregoal});
	\item evaluate $\bm{J} \triangleq \bm{J}^{c} = 0.5 \left( \bm{J}^{e} + \bm{J}^{g} \right)$.
\end{enumerate}
These choices, however, provide unsatisfactory Cartesian trajectory of the end-effector. A pictorial view of the resulting trajectories is shown in Fig. \ref{fig:trajectories}.
As a result, we designed a new image-based visual servoing control that provides satisfactory trajectories and complies with requirements $\left( \textsc{i} \right)$ and $\left( \textsc{ii} \right)$.

Our approach considers two image-based visual servoing problems to be solved.
The first solves for the translation motion assuming the rotation completed.
This is equivalent to consider the current pose of the end-effector as the combination of the 3D Cartesian component of $\bm{x}^{e}$ and the axis-angle representation of the orientation of $\bm{x}^{g}$, i.e.
\begin{equation}
	\bm{x}_{t}^{e} \triangleq \left[ p_{x}^{e}, p_{y}^{e}, p_{z}^{e}, u_{x}^{g}, u_{y}^{g}, u_{z}^{g}, \theta^{g} \right]^{\top} .
\end{equation}
Conversely, in the second problem we compute the rotation motion under the assumption of achieved translation, i.e.
\begin{equation}
	\bm{x}_{o}^{e} \triangleq \left[ p_{x}^{g}, p_{y}^{g}, p_{z}^{g}, u_{x}^{e}, u_{y}^{e}, u_{z}^{e}, \theta^{e}\right]^{\top} .
\end{equation}
We then proceed with the classic approach, defining four coplanar 3D points around $\bm{x}_{t}^{e}$ and $\bm{x}_{o}^{e}$ as in (\ref{eq:featureend}), i.e.
\begin{equation}
	\bm{\Pi}_{l} \bm{x}^{e}_{t, i} = z^{e}_{l,t,i} \!
	\left[
		\begin{IEEEeqnarraybox*}[][c]{,c,}
			u^{e}_{l,t,i} \vspace{0.5em}\\
			v^{e}_{l,t,i} \vspace{0.5em}\\
			1
		\end{IEEEeqnarraybox*}
	\right]
	\!,\,
	\bm{\Pi}_{r} \bm{x}^{e}_{t, i} = z^{e}_{r,t,i} \!
	\left[
		\begin{IEEEeqnarraybox*}[][c]{,c,}
			u^{e}_{r,t,i} \vspace{0.5em}\\
			v^{e}_{r,t,i} \vspace{0.5em}\\
			1
		\end{IEEEeqnarraybox*}
	\right]
	\!,\,
	1 \leq i \leq 4 \label{eq:featureendt}
\end{equation}
\begin{equation}
	\bm{\Pi}_{l} \bm{x}^{e}_{o, i} = z^{e}_{l,o,i} \!
	\left[
		\begin{IEEEeqnarraybox*}[][c]{,c,}
			u^{e}_{l,o,i} \vspace{0.5em}\\
			v^{e}_{l,o,i} \vspace{0.5em}\\
			1
		\end{IEEEeqnarraybox*}
	\right]
	\!\!,
	\bm{\Pi}_{r} \bm{x}^{e}_{o, i} = z^{e}_{r,o,i} \!
	\left[
		\begin{IEEEeqnarraybox*}[][c]{,c,}
			u^{e}_{r,o,i} \vspace{0.5em}\\
			v^{e}_{r,o,i} \vspace{0.5em}\\
			1
		\end{IEEEeqnarraybox*}
	\right]
	\!\!,
	1 \leq i \leq 4 \label{eq:featureendo}
\end{equation}
the visual features $\bm{s}_{t}^{e}$ and $\bm{s}_{o}^{e}$ as in (\ref{eq:featee}), two image Jacobians $\bm{J}_{t}^{e}$ and $\bm{J}_{o}^{e}$ using, respectively, (\ref{eq:featureendt}) and (\ref{eq:featureendo}) as in (\ref{eq:jacobian}), and finally the error functions
\begin{IEEEeqnarray}{rCl}
	\bm{e}_{t} & \triangleq & \bm{s}_{t}^{e} - \bm{s}^{g} \, , \label{eq:visualservoerrort}\\
	\bm{e}_{o} & \triangleq & \bm{s}_{o}^{e} - \bm{s}^{g} \, . \label{eq:visualservoerroro}
\end{IEEEeqnarray}
The velocity screws
\begin{IEEEeqnarray}{rCl}
	\dot{\bm{x}}^{e}_{t} & \triangleq & \left[ \, \bm{v}_{t},\, \bm{\omega}_{t} \, \right]^{\top} ,\, \label{eq:velocityscrewt}\\
	\dot{\bm{x}}^{e}_{o} & \triangleq & \left[ \, \bm{v}_{o},\, \bm{\omega}_{o} \, \right]^{\top} ,\, \label{eq:velocityscrewo}
\end{IEEEeqnarray}
are computed with
\begin{IEEEeqnarray}{rCl}
	\dot{\bm{x}}^{e}_{t} & = & -K^{e}_{t} \left( \bm{J}^{e}_{t} \right)^{\dagger} \bm{e}_{t} \, , \label{eq:controlvelt}\\
	\dot{\bm{x}}^{e}_{o} & = & -K^{e}_{o} \left( \bm{J}^{e}_{o} \right)^{\dagger} \bm{e}_{o} \, . \label{eq:controlvelo}
\end{IEEEeqnarray}
Finally, (\ref{eq:velocityscrewt}) and (\ref{eq:velocityscrewo}) are combined in a single velocity screw $\dot{\bm{x}}^{e} = \left[ \, \bm{v}_{t},\, \bm{\omega}_{o} \, \right]^{\top}$ that is used by the Cartesian controller to command the robot's end effector.
The resulting trajectory turns out to be satisfactory, combining a decoupled translation and rotation motion.
A comparison view of the trajectories is shown in Fig. \ref{fig:trajectories}.

\subsection{Gain scheduling}
\label{ssec:gainscheduling}
The choice of $K^{e}$ in (\ref{eq:controlvel}) is critical because it affects both the speed of the final movement and the convergence to the goal pose.
On the one hand, high $K^{e}$ may lead to overshooting and/or oscillating around the goal, on the other hand, a low $K^{e}$ would increase the control convergence, but the resulting movement will be slow.
A practical, yet effective, solution to the above-mentioned considerations is to have different gains depending on the robot's operative point.

In this work, we use a \textit{gain-scheduling} approach \cite{Astrom:2013aa} to change the value of $K^{e}$ when the end-effector is close to the goal pose.
As a result, the control law (\ref{eq:controlvel}) becomes as follows:
\begin{IEEEeqnarray}{rCl}
	\dot{\bm{x}}^{e} & = & \left \{
	\begin{IEEEeqnarraybox*}[][c]{;c;l;}
		- K^{e}_{1} J^{\dagger} \bm{e} & \mbox{, if } ||\bm{e}|| \ge \tau_{e} \vspace{0.5em} \\
		- K^{e}_{2} J^{\dagger} \bm{e} & \mbox{, otherwise}
	\end{IEEEeqnarraybox*}
	\right. \, , \label{eq:gainscheduling}
\end{IEEEeqnarray}
where $\tau_{e}$ is a distance threshold and $K^{e}_{1} \ge K^{e}_{2} > 0$ are the two proportional gains.


\begin{figure*}[t!]
\centering
	\framebox{\parbox{0.27\linewidth}{\includegraphics[width=\linewidth]{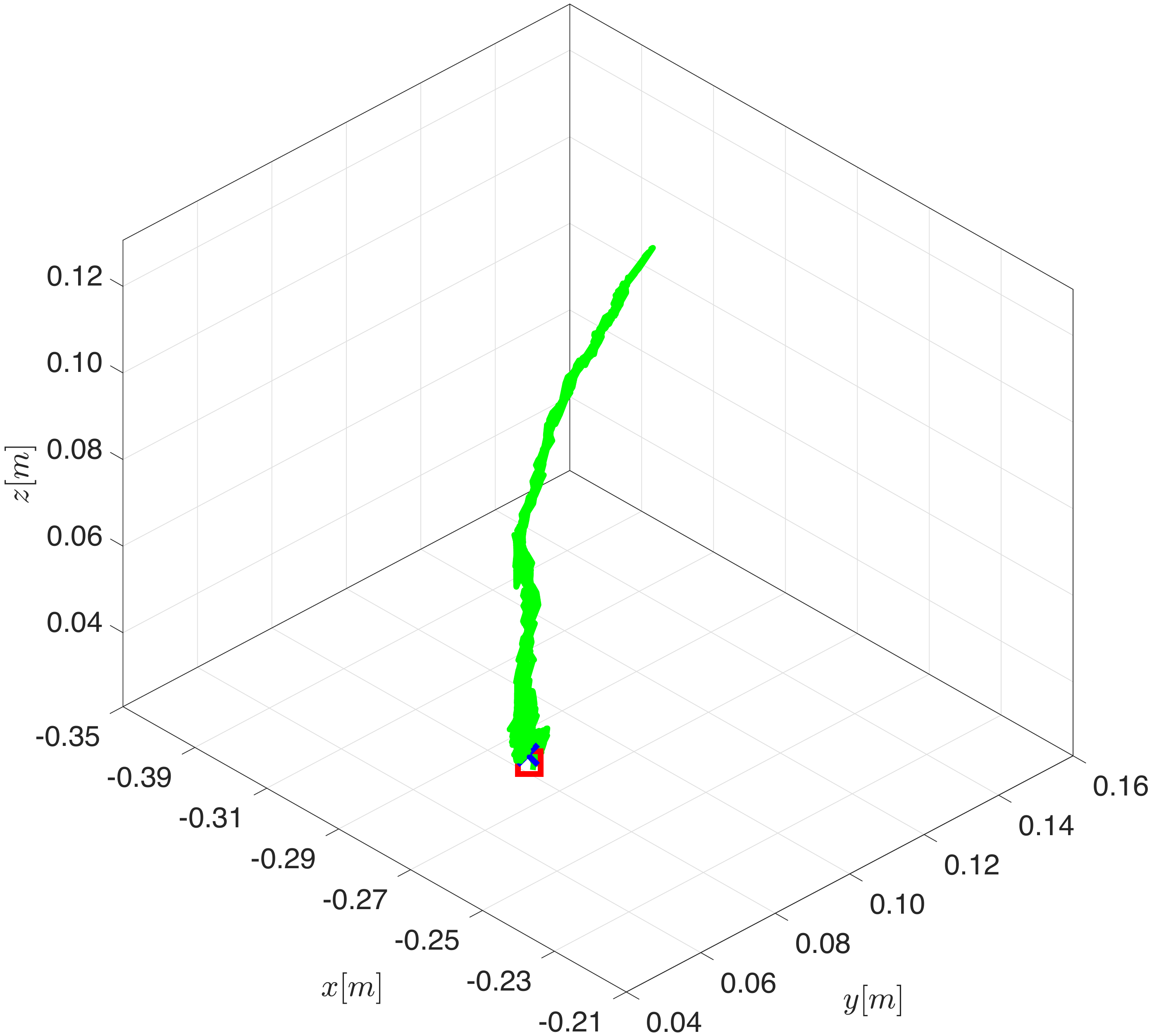}}}
	\framebox{\parbox{0.325\linewidth}{\includegraphics[width=\linewidth]{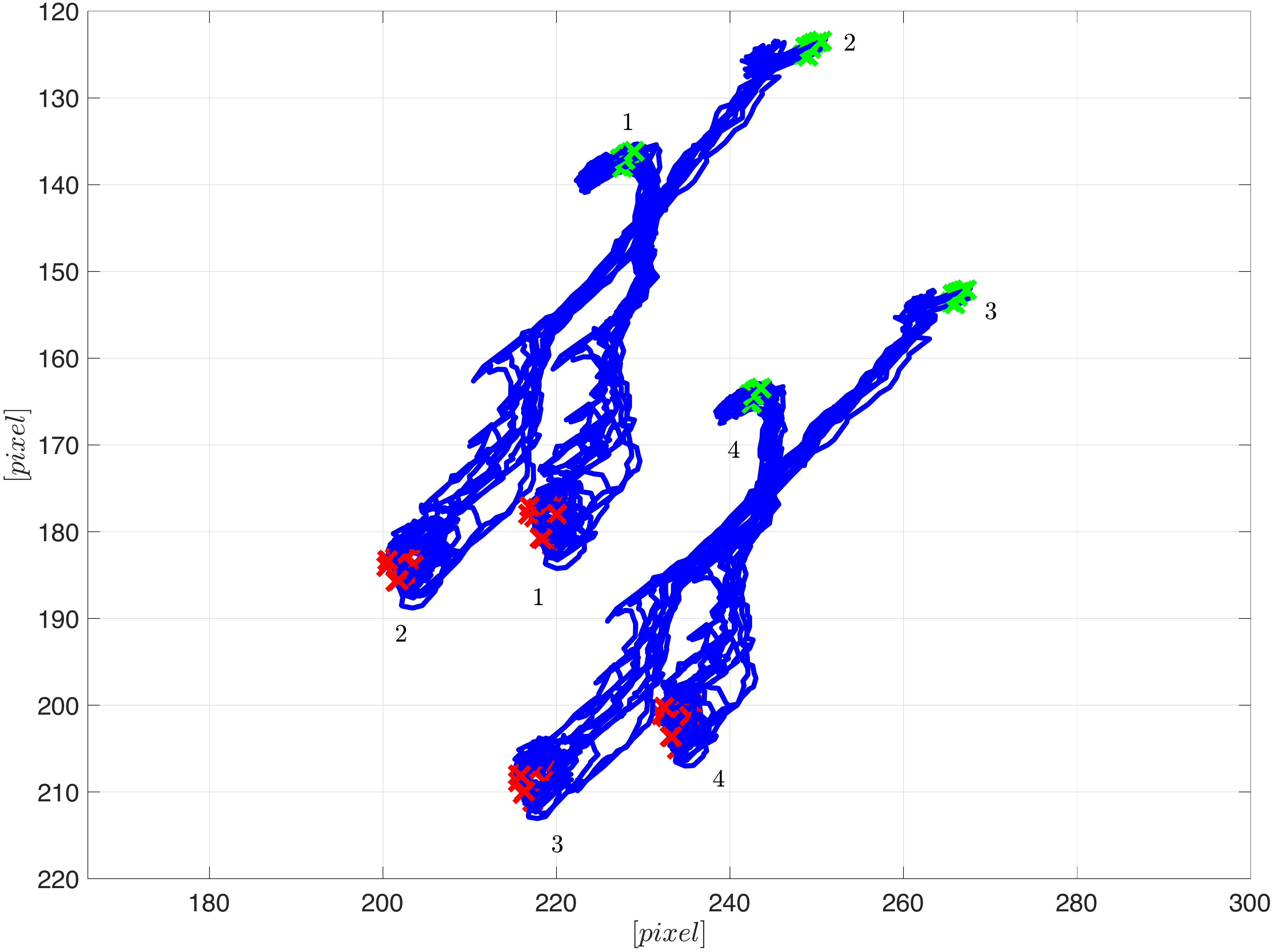}}}
	\framebox{\parbox{0.323\linewidth}{\includegraphics[width=\linewidth]{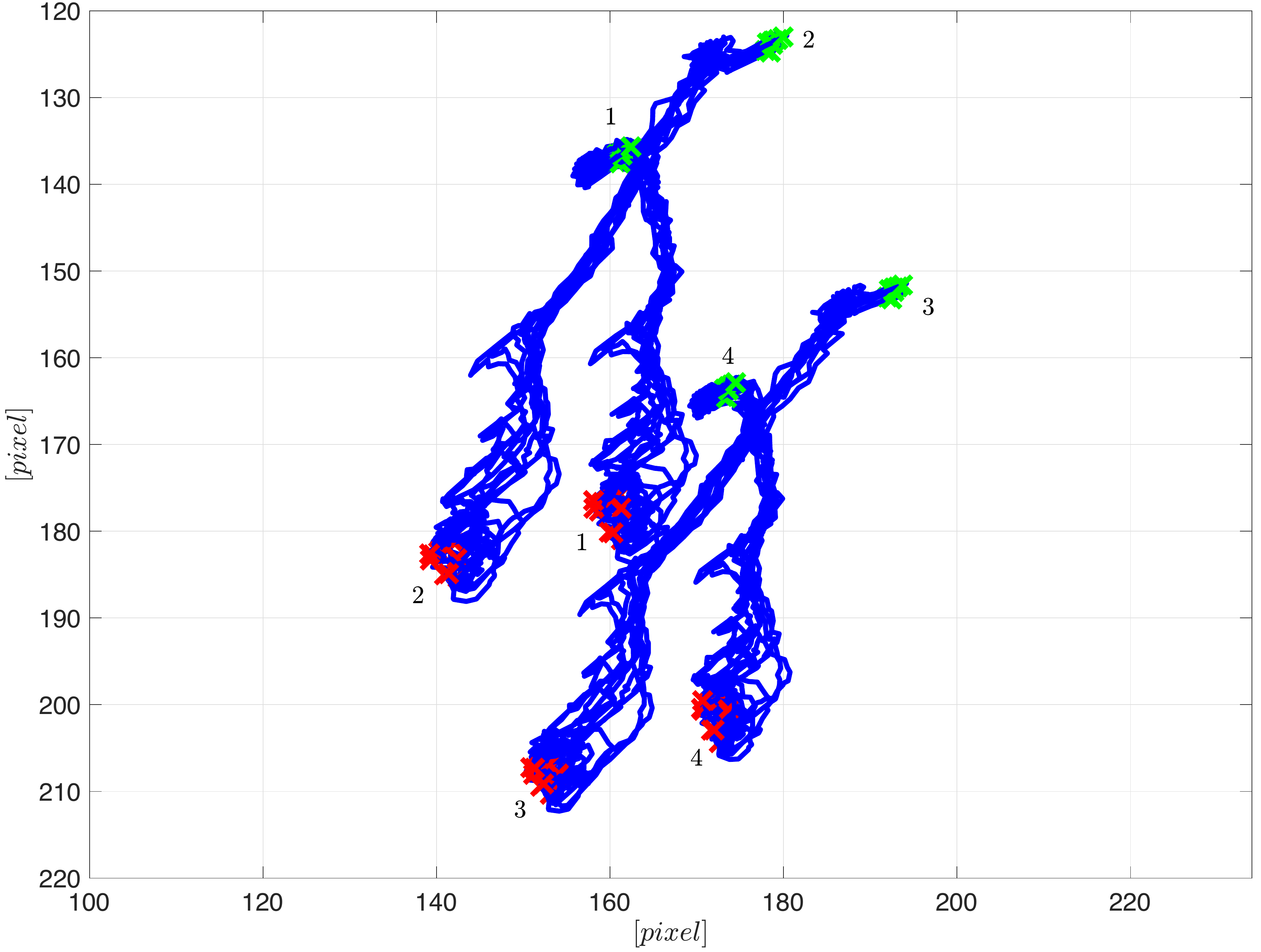}}}
        \caption{\textbf{Left:} 3D Cartesian trajectories of the right end-effector. \textbf{Center/Right:} Image coordinates on the left/right camera of the four points representing the pose of the right end-effector. Green/Red crosses represents the starting/goal positions. Note that the starting position numbers $1$-$4$ are ordered clockwise, while the goal positions numbers are ordered counter-clockwise, implying for a $180 \, \left[^{\circ} \right]$ rotation.}
        \label{fig:exp2}
        \vspace{0.5em}
\hrulefill
\vspace{-1em}
\end{figure*}

\section{EXPERIMENTAL RESULTS}
\label{sec:exp}
To evaluate the effectiveness and robustness of the proposed framework, a C++ implementation of the pipeline has been tested on the iCub humanoid robot platform.
We ran our experiments on two laptops, shipped with an Intel i7 processor, and a workstation equipped with a NVIDIA K40 GPU in order to use the CUDA \cite{cuda} HOG implementation provided by OpenCV \cite{opencv}.

\begin{figure}[thpb]
\centering
	\framebox{\parbox{0.27\linewidth}{\includegraphics[width=\linewidth]{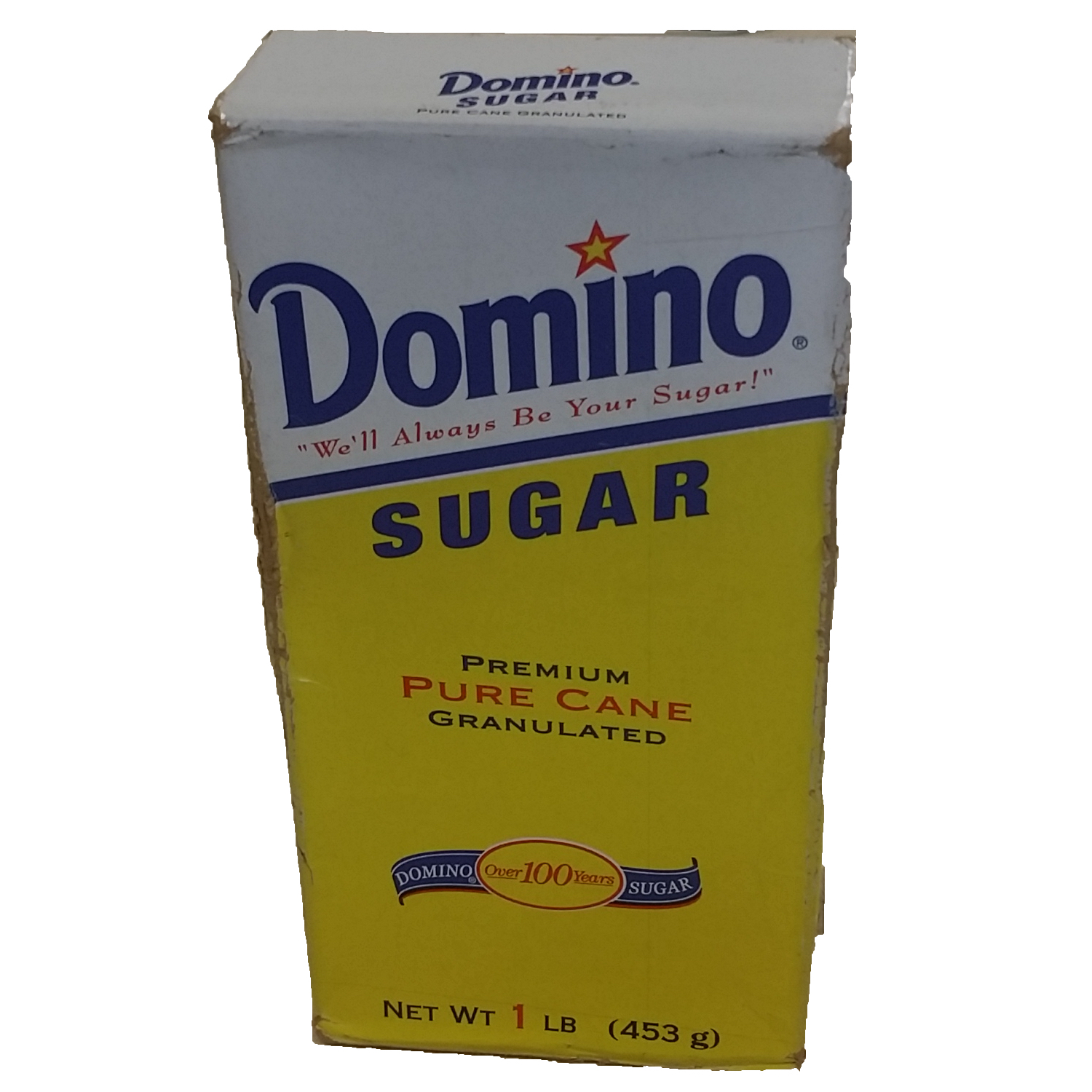}}}
	\framebox{\parbox{0.27\linewidth}{\includegraphics[width=\linewidth]{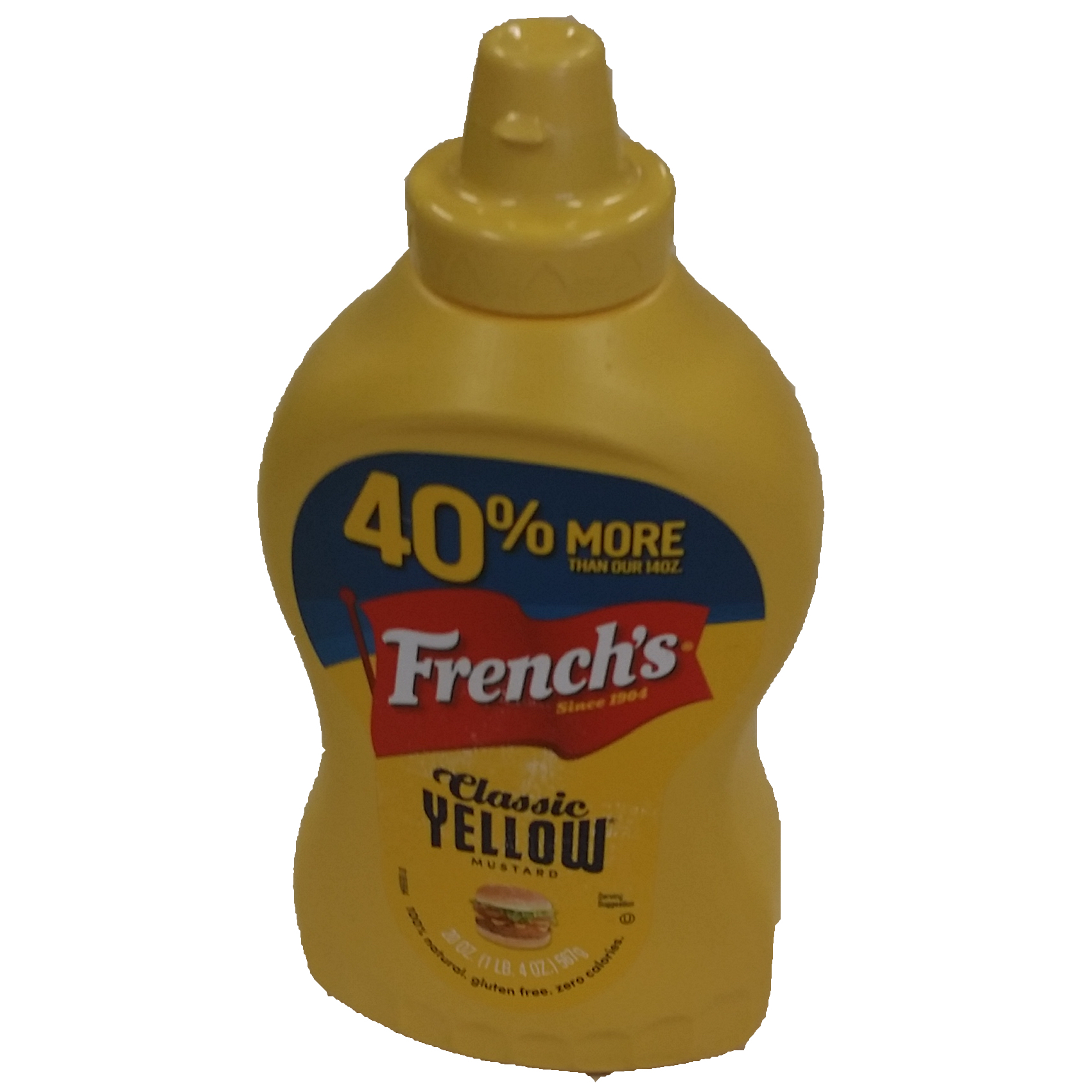}}}
	\framebox{\parbox{0.27\linewidth}{\includegraphics[width=\linewidth]{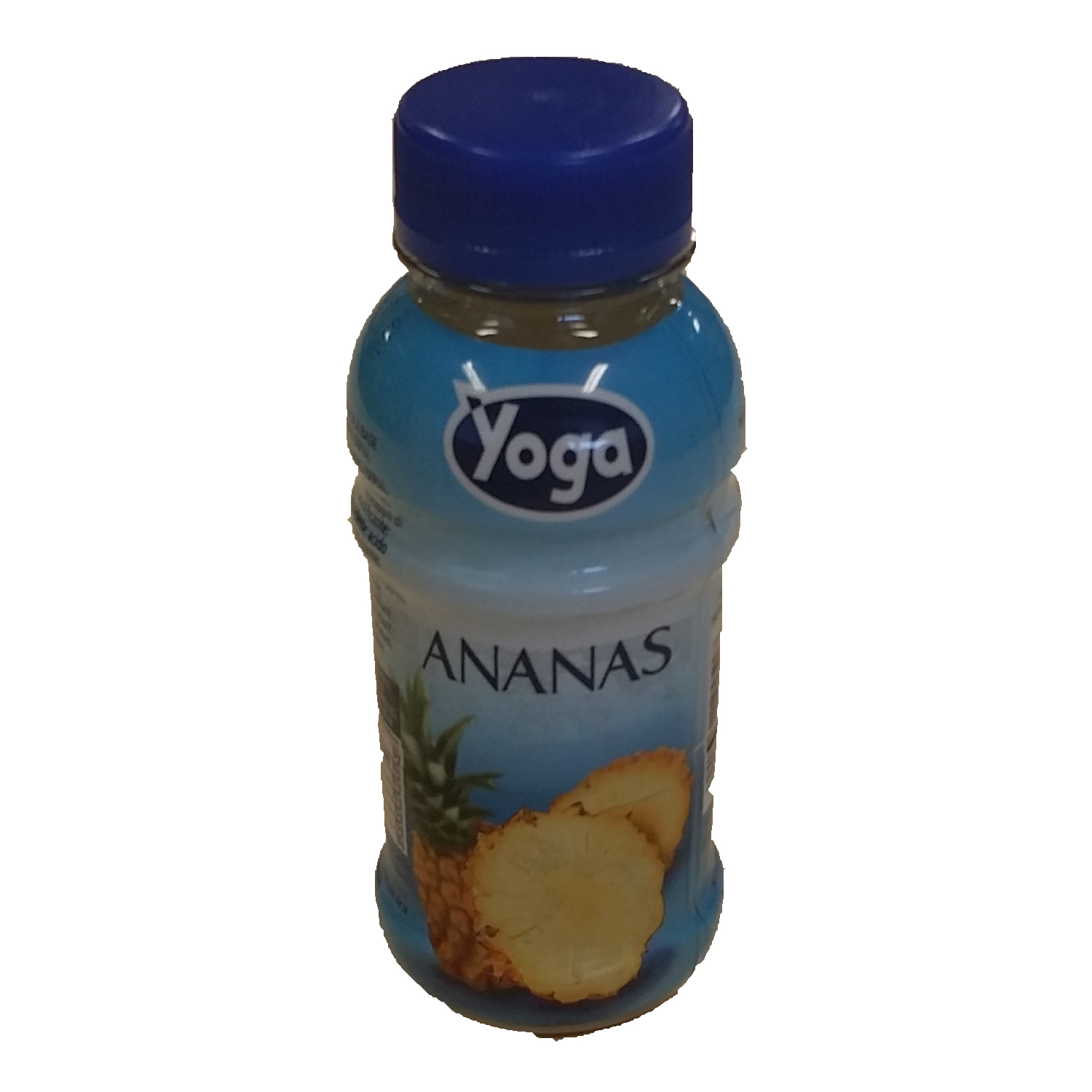}}}
        \caption{\textbf{Left:} Domino sugar box of the YCB dataset \cite{ycb}. \textbf{Center:} French's mustard bottle of the YCB dataset \cite{ycb}. \textbf{Right:} Pineapple juice bottle, a small everyday cylinder-shaped object.}
        \label{fig:objects}
\end{figure}

We carried out two different experiments: $\left. 1 \right)$ 10 grasps on 3 different objects and $\left. 2 \right)$ 10 reaching motions using the same initial and goal pose.
On the one hand, the goal of the first experiment is to assess the robustness of the pipeline as a whole.
The 3 objects have different shape and size, see Fig. \ref{fig:objects}, in order to put to test the superquadric modeling, the grasping pose computation and the precision of the visual servoing control.
On the other hand, the second experiment focuses on assessing the repeatability of the generated trajectories.
In particular, we tested the 3D model aided particle filter and the image-based visual servo control by performing a trajectory that involves a translation and $180 \, [^{\circ}]$ rotation.

Both experiment 1 and 2 use the same set of parameters.
Details of the grasping and 3D model-aided particle filter implementation can be found on Github\footnote{\faGithub\,\textit{github.com/robotology/superquadric-model/tree/master} \\ \hbox{~~~~}\faGithub\,\textit{github.com/robotology/superquadric-grasp/tree/feature-visualservoing} \\ \hbox{~~~~}\faGithub\,\textit{github.com/robotology/visual-tracking-control/tree/master}}.
The visual servoing gains with gain scheduling are $K^{e}_{t, 1} = 0.5$, $K^{e}_{t, 2} = 0.25$, $K^{e}_{o, 1} = 3.5$ and $K^{e}_{o, 2} = 0.5$, with distance thresholds $\tau^{e}_{t} = \tau^{e}_{o} = 10 \, \left[ pixel \right]$.
The termination condition for visual servoing is achieved when the $\ell^{2}$-norm of $e$ falls below $1$ pixel.

\subsection{Experiment 1}
To evaluate the performance of experiment 1, we calculate, for each object, the \textit{Root Mean Square Error} (RMSE) of both the image and Cartesian coordinates of the end-effector pose.
Further, in order to have a good term of comparison, we also calculate the RMSE of the image and Cartesian coordinates that would have been obtained if the direct kinematics would have been used in place of the estimates of the particle filter.
We term the RMSE of the image coordinates \textit{Image RMSE} (IRMSE), the Cartesian position error \textit{Position RMSE} (PRMSE) and the orientation error \textit{Orientation RMSE} (ORMSE).

Table \ref{tab:exp1} reports the RMSE obtained with 10 grasping.
As it can be seen, the IRMSE decreases by three order of magnitude and we successfully achieved sub-pixel precision.
The PRMSE is decreased by an order of magnitude, achieving millimeter precision, and the ORMSE by almost a factor of 2.
\begin{table}[thpb]
	\vspace{-1em}
	\renewcommand{\arraystretch}{1.3}
	\caption{RMSE of experiment 1}
	\label{tab:exp1}
	\centering
	\begin{tabular}{cccc}
		Name & IRMSE $[pixel]$ & PRMSE $[m]$ & ORMSE $[^{\circ}]$\\
		\hline\hline
		& $0.857$ & $0.004$ & $2.071$ \vspace{-0.85em}\\
		Sugar box &&& \vspace{-0.85em}\\
		& $13.171$ & $0.028$ & $7.482$\\
		\hline
		& $0.871$ & $0.003$ & $2.46$ \vspace{-0.85em}\\
		Mustard bottle &&& \vspace{-0.85em}\\
		& $11.491$ & $0.021$ & $6.098$\\
		\hline
		& $0.887$ & $0.003$ & $3.08$ \vspace{-0.85em}\\
		Juice bottle &&& \vspace{-0.85em}\\
		& $11.202$ & $0.022$ & $5.73$\\
		\hline\hline
	\end{tabular}
\end{table}

\subsection{Experiment 2}
The iCub is required to start form the pose $\bm{x}^{e} = \left[ -0.28, 0.12, 0.13, 0.131, -0.492, 0.86, 2.962 \right]^{\top}$, with the palm oriented upward, and to reach the pose $\bm{x}^{g} = \left[ -0.28, 0.08, 0.03, 0.213, -0.94, 0.265, 2.911 \right]^{\top}$, with the palm oriented downward, for 10 times.
No objects are present in this setting, thus the final goal is provided manually and it is thus not used to evaluate the pose error.
Instead, we are interested in assessing whether or not the trajectories are smooth, with decoupled translation and orientation motion, and reproducible given the same initial and final pose.
The termination condition is achieved when the $\ell^{2}$-norm of $\bm{e}$ falls below $1 \, [pixel]$.

The robot achieved sub-pixel precision for all $10$ trials with an IRMSE of $0.855 \, [pixel]$
A pictorial view of the trajectories are shown in Fig. \ref{fig:exp2}
Note that our framework has the capability of providing smooth and reproducible trajectories, with the desired behaviour of decoupled translation and rotation motion.


\section{CONCLUSIONS AND FUTURE WORK}
\label{sec:con}
This paper proposed a new framework for markerless visual servoing on unknown objects which consists of the following methods: a grasping approach for estimating the 3D shape, pose and grasping pose of unknown objects using stereo information; a Bayesian recursive filtering approach, based on Sequential Monte Carlo filtering, for estimating the pose of the robot's end-effector using RGB images; a novel image-based visual servoing approach capable of providing decoupled translation and orientation control.

It was shown, through experimental results on the iCub humanoid robot platform, that our framework is robust and reliable, providing a significant improvement in terms of Root Mean Square Error in both image and Cartesian coordinates with respect to using information provided by the direct kinematics.
Further we also showed that given the same initial and goal position, we can achieve smooth trajectories with decoupled translation and rotation motion.

The objects taken into account during the experimental evaluation favored lateral grasps due to their elongated shape.
However, smaller  and differently shaped objects might be better grasped from the top or with grasping poses that partially occlude the hand.
In this scenario, our framework, and in particular the end-effector pose estimation, can be easily extended to include the forearm CAD model, thus accounting for partial or complete occlusion of the hand.
This is toward testing our framework on a larger set of objects, like the YCB dataset \cite{ycb}.





\bibliographystyle{ieeetr}
\bibliography{IEEEabrv,icra_2018}

\end{document}